\documentclass[11pt]{article}
\usepackage{graphicx}
\usepackage[shortlabels]{enumitem}
\usepackage{amsmath,amsthm}
\usepackage{amsfonts}
\usepackage{mathtools}
\usepackage{xcolor}
\usepackage{booktabs}
\usepackage{todonotes}
\usepackage{jlcode}
\usepackage{float}
\usepackage{subcaption}
\usepackage{hyperref}

\usepackage{siunitx}

\usepackage[commentmarkup=uwave]{changes}
\definechangesauthor[color=orange]{BL}

\newtheorem{theorem}{Theorem}

\title{A General and Streamlined \\ Differentiable Optimization Framework}
\author{Andrew~W.~Rosemberg,
\; Joaquim Dias Garcia, \\
\; François Pacaud,
\; Robert B. Parker,
\; Benoît Legat, \\
\; Kaarthik Sundar,
\; Russell Bent,
\; Pascal Van Hentenryck
\footnote{Andrew~W.~Rosemberg and Pascal Van Hentenryck are with the Georgia Institute of Technology; Joaquim Dias Garcia is with PSR; François Pacaud is with Mines Paris-PSL; Robert B. Parker, Kaarthik Sundar and Russell Bent are with the Los Alamos National Laboratory; and Benoît Legat is with UCLouvain.}
}

\date{August 2025}

\makeatletter
\DeclareRobustCommand{\cev}[1]{%
  {\mathpalette\do@cev{#1}}%
}
\newcommand{\do@cev}[2]{%
  \vbox{\offinterlineskip
    \sbox\z@{$\m@th#1 x$}%
    \ialign{##\cr
      \hidewidth\reflectbox{$\m@th#1\vec{}\mkern4mu$}\hidewidth\cr
      \noalign{\kern-\ht\z@}
      $\m@th#1#2$\cr
    }%
  }%
}
\makeatother

\begin{document}

\maketitle

\begin{abstract}
Differentiating through constrained optimization problems is increasingly central to learning, control, and large-scale decision systems, yet practical integration remains difficult due to solver specialization and interface mismatch. This paper presents a general and streamlined framework—an updated \texttt{DiffOpt.jl}—that unifies modeling and differentiation within the Julia optimization stack. The framework computes forward- and reverse-mode \emph{solution} and \emph{objective} sensitivities for smooth, potentially nonconvex programs by differentiating the KKT system under standard regularity assumptions. A first-class, \texttt{JuMP}-native \emph{parameter-centric} API allows users to declare named parameters and obtain derivatives directly with respect to them—even when a parameter appears in multiple constraints and objectives—eliminating brittle bookkeeping from coefficient-level interfaces.

We illustrate these capabilities on convex and nonconvex models, including economic dispatch, mean–variance portfolio selection with conic risk constraints, and nonlinear robot inverse kinematics. Two companion studies further demonstrate impact at scale: gradient-based iterative methods for strategic bidding in energy markets and Sobolev-style training of end-to-end optimization proxies using solver-accurate sensitivities. Together, these results show that differentiable optimization can be deployed as a routine tool for experimentation, learning, calibration, and design—without departing from standard \texttt{JuMP} modeling practice and while retaining access to a broad ecosystem of solvers.
\end{abstract}

\section{Acknowledgments}
This research is partly funded by NSF award 2112533. 
The work was also funded by Los Alamos National Laboratory's Directed Research and Development project, ``Artificial Intelligence for Mission (ArtIMis)'' under U.S. DOE Contract No. DE-AC52-06NA25396.
LA-UR-25-30199.

\section{Introduction}
\label{sec:intro}

Understanding how the solutions of constrained optimization problems vary with respect to parameters has become a core objective in many modern applications. From estimating the impact of noisy data on problem outcomes to performing end-to-end training that embeds decision-making layers within deep learning architectures, the ability to differentiate through constrained optimization can greatly enhance the performance of real-world systems.
The growing popularity of \emph{differentiable layers} that encapsulate entire decision problems underscores the value of computing solution sensitivities and gradients, thereby allowing models to optimize \emph{decision quality} rather than solely fitting predictive objectives.

A rich body of work has studied solution sensitivity and the role of derivatives in constrained optimization pipelines. Early research by \cite{fiacco1976sensitivity} established conditions---such as uniqueness of primal and dual solutions and stability of the active set---under which the implicit function theorem describes how optimal solutions shift under parameter perturbations. Subsequent efforts have relaxed these assumptions \cite{kojima1980,jittorntrum1984,shapiro1985,ralph1995}, extended analysis to broader contexts (e.g., bi-level programs \cite{dempe2002} or stochastic formulations \cite{shapiro1990,shapiro1991,zavala2009,jin2014}), and implemented practical software for this purpose \cite{pirnay2012optimal,andersson2019casadi}. In parallel, advances in \emph{differentiable programming} \cite{innes2019differentiable} have greatly facilitated automatic gradient computations across broad classes of programs, providing an appealing way to unify optimization-based modules with gradient-based machine learning pipelines \cite{amos2017optnet,agrawal2019differentiable,pineda2022theseus}.

Despite these theoretical and algorithmic advances, integrating differentiable constrained optimization into large-scale industrial pipelines or research prototypes remains nontrivial. 
Practitioners still face persistent barriers when trying to use sensitivity information at scale as many tools are specialized to narrow classes (e.g., convex QP/SOCP only) or require nonstandard modeling abstractions (e.g., the \texttt{suffix} data structures that must be populated to use sIPOPT \cite{pirnay2012optimal,sipoptmanual}).
In response to this gap, \texttt{DiffOpt.jl} \cite{besanccon2024flexible} was developed to provide a Julia framework for differentiating through the solutions of optimization problems. By building on \texttt{MathOptInterface.jl} \cite{lubin2023jump}, \texttt{DiffOpt.jl} seamlessly leverages a large set of solvers as \texttt{MathOptInterface.jl} automatically applies solver-compatible transformations so that the user can remain focused on modeling rather than solver internals. 
However, three key limitations remained:
\begin{itemize}
    \item \texttt{DiffOpt.jl} primarily addressed convex optimization problems, while real applications often involve non-convex, continuous problems (e.g., certain classes of nonlinear constraints or objectives) for which existing differentiable techniques and implementations are far less mature.
    \item \texttt{DiffOpt.jl} takes a function-centric view of sensitivity, returning derivatives with respect to the coefficients of each function in the objective or constraints. While this is technically general, users often care about \emph{explicit problem parameters} that may appear in multiple places (e.g., repeated in several constraints or in both constraints and objective terms). Mapping from function-based derivatives to parameter-based derivatives adds complexity and potential confusion.
    \item The interface in the style of \texttt{MathOptInterface.jl} enables the user to have access to the required model transformations, but it is still not as streamlined as the \texttt{JuMP.jl} interface -- a high-level, user-friendly modeling abstraction on top of \texttt{MathOptInterface.jl} that provides a simple syntax mirroring standard mathematical notation.
\end{itemize}

This paper presents an updated version of \texttt{DiffOpt.jl} that closes these gaps. The new implementation computes sensitivities for broad classes of \emph{continuous, potentially nonconvex} problems and lets users query derivatives directly with respect to named, explicit parameters. The latter change removes the need for brittle bookkeeping when one parameter influences many constraints and makes gradients directly consumable by training loops, calibration routines, and robustness analyses. Extending support to nonconvex solvers brings sensitivities to settings where convex approximations are unavailable or too coarse—including robotics and mechanism design, scientific ML and control with physics-based constraints, engineering design with nonlinear performance limits, portfolio and risk models with conic–nonlinear couplings, and networked systems (e.g., gas and power) with nonconvex flows. Exposing these capabilities through a \texttt{JuMP}-native interface lowers adoption costs: existing models can be retained, parameters declared, and differentiation invoked without departing from standard modeling practice. 

Two companion studies illustrate how the capabilities introduced here enable new results at scale. First, \emph{Strategic Bidding in Energy Markets with Gradient-Based Iterative Methods} \cite{rosemberg2025strategic} demonstrates how iterative methods for solving bilevel problems leveraging solution sensitivities of lower level optimization problems dramatically improve both solution quality and computational efficiency. In particular, the mentioned paper depends crucially on the ability to compute sensitivities for nonconvex programs. 
With our updates to \texttt{DiffOpt.jl}, this can be accomplished while leveraging existing \texttt{JuMP} models. 

Second, \emph{Sobolev Training of End-to-End Optimization Proxies} \cite{rosemberg2025sobolev} uses solver-based sensitivities to train fast surrogates for large optimization workloads while preserving correctness of derivatives. The study shows substantial gains in accuracy and feasibility when supervising proxies with masked Jacobian information, and it relies on differentiating the solution map with respect to explicit problem parameters to construct the training signal. These experiments—and their scale—are made feasible by the parameter-centric interface and the efficient KKT-based reverse mode implemented in the present framework.

Our contributions are therefore not a new theory of sensitivity, but a unification of modeling and differentiation that makes established theory \emph{usable} across problem classes and application domains. We show how to compute solution and objective sensitivities in forward and reverse mode for general continuous problems, how to expose derivatives with respect to named parameters without forcing users into function-coefficient coordinates, and how to do so while leveraging solver back-ends and model transformations in the Julia optimization ecosystem \cite{bezanson2017julia}. The result is a framework that turns differentiable optimization from a boutique capability into a routine tool for large-scale experimentation, learning, calibration, and design.

The remainder of the paper reviews the technical background on parametric optimization and KKT-based differentiation, describes the architecture and interface that realize the proposed capabilities, and reports case studies spanning convex and nonconvex models. Throughout, the emphasis is on clarity of modeling, fidelity of sensitivities, and ease of integration with modern computational workflows.

\section*{Literature Review}

\subsection*{Sensitivity Analysis and the Implicit Function Theorem}
Classical work by \cite{fiacco1976sensitivity} established conditions under which the KKT equations of a constrained optimization problem admit stable, differentiable solutions. The key requirements include:
\begin{enumerate}
    \item The \emph{primal solution mapping} is single-valued (unique solution),
    \item The \emph{dual solution} is unique,
    \item The \emph{active set} is locally stable.
\end{enumerate}
Under second-order sufficiency (SOSC), linear independence constraint qualification (LICQ), and strict complementarity slackness (SCS), the KKT system can be viewed as a set of smooth nonlinear equations in terms of the decision variables and parameters. The \emph{implicit function theorem} then enables us to compute how optimal solutions vary with respect to parameter changes.

Subsequent work relaxed these conditions to handle degeneracies. For instance, \cite{kojima1980} introduced Mangasarian Fromovitz constraint qualification (MFCQ) and the generalized strong second order sufficient condition (GSSOSC) to ensure strong stability without requiring strict complementarity (though differentiability may fail in such cases). In \cite{jittorntrum1984} and \cite{shapiro1985}, the authors studied directional differentiability of solutions even without LICQ. Practical methods for evaluating directional derivatives were explored in \cite{ralph1995}. In the following decades, the scope of sensitivity analysis expanded further to settings like variational inequalities, bi-level programming \cite{dempe2002}, stochastic optimization \cite{shapiro1990,shapiro1991}, and model predictive control \cite{zavala2009,jin2014}, culminating in efficient software implementations \cite{pirnay2012optimal,andersson2019casadi}.

\subsection*{Differentiable Programming and Optimization Layers}
Since the 2010s, the fields of gradient-based optimization \cite{huangfu2018parallelizing,shin2023accelerating,lubin2023jump} and differentiable programming \cite{innes2019differentiable} have grown rapidly. Integrating constrained optimization into neural networks has proven particularly useful in machine learning, facilitating hyperparameter tuning, end-to-end learning for decision pipelines, and other applications \cite{amos2017optnet}.
Software libraries like CVXPY-Layers \cite{agrawal2019differentiable}, sIPOPT \cite{pirnay2012optimal}, CasADi \cite{andersson2019casadi}, and Theseus \cite{pineda2022theseus} provide toolkits for embedding optimization within differentiable models. More recently, \texttt{DiffOpt.jl} \cite{besanccon2024flexible} extended \texttt{JuMP.jl} in the Julia ecosystem to compute solution sensitivities for convex problem classes. By integrating these capabilities with standard deep learning frameworks (e.g., \texttt{Pytorch}, \texttt{Tensorflow} and \texttt{Flux.jl}), one can form \emph{differentiable layers} that solve optimization problems given the outputs of preceding network layers.

\section{Technical Background}
\label{sec:background}
\subsection{Parametric Optimization}
Parametric optimization problems arise in scenarios where certain elements (e.g., coefficients, constraints) may vary according to problem parameters. A general form of a parameterized non-convex optimization problem is
\begin{equation}
  \label{eq:problem}
\begin{aligned}
    \min_{x} \; & \; f(x ; p) \\
    \; \text{s.t.}\; & \; c(x; p) = 0 \;, \; x \geq 0 \; .
\end{aligned}
\end{equation}
We note by $x \in \mathbb{R}^n $ the optimization variable, while the problem’s structure is encoded by the objective
function $ f : \mathbb{R}^n \rightarrow \mathbb{R} $ and the equality constraints $ c : \mathbb{R}^n \rightarrow \mathbb{R}^{m} $. The problem \eqref{eq:problem} depends on a parameter $p \in \mathbb{R}^{\ell}$
appearing in the objective and the constraints.
We suppose that the two functions $f(\cdot; \cdot)$ and $c(\cdot; \cdot)$ are smooth and twice-differentiable.

\paragraph{Karush-Kuhn-Tucker (KKT) optimality conditions.}
In \eqref{eq:problem}, we note $\lambda \in \mathbb{R}^m$ the Lagrange multiplier associated
to the equality constraints and $\nu \in \mathbb{R}^n$ the multiplier associated
to the bound constraints. We define the Lagrangian of \eqref{eq:problem} as:
\begin{equation}
  \label{eq:lagrangian}
  L(x, \lambda, \nu; p) = f(x; p) + \lambda^\top c(x; p) - \nu^\top x \; .
\end{equation}
The KKT stationary conditions associated to the problem~\eqref{eq:problem} are:
\begin{equation}
  \label{eq:kkt}
  \left\{
  \begin{aligned}
    & \nabla_x f(x; p) + \nabla_x c(x; p)^\top \lambda - \nu = 0 \\
    & c(x; p) = 0 \\
    & 0 \leq x \perp \nu \geq 0 \; .
  \end{aligned}
  \right.
\end{equation}
The notation $0 \leq x \perp \nu \geq 0$ is used to denote the complementarity
condition $x_i \nu_i = 0$, $x_i \geq 0$, $\nu_i\geq 0$ for $i = 1, \cdots, n$.

\subsection{Sensitivity Calculations}

Since the seminal work of Fiacco~\cite{fiacco1976sensitivity},
the calculation of sensitivity usually proceeds by rewriting necessary KKT conditions \eqref{eq:kkt} as a system of nonlinear equations that locally define the optimal primal-dual solution.
Then, the derivative follows from the Implicit Function Theorem (IFT) \cite{dontchev2009}.
However, this method requires the problem~\eqref{eq:problem} to be regular enough
so that the active set in the constraint $x \geq 0$ does not change close to the solution.

This section will outline the derivations of the optimality conditions for a
general optimization problem and apply the Implicit Function Theorem to derive the sensitivities.

\subsection{Sensitivity and Implicit Function Theorem}

Following \cite{dontchev2009}, the next theorem gives a statement for the IFT.

\begin{theorem}
Let $ F : \mathbb{R}^{n} \times \mathbb{R}^{\ell} \rightarrow \mathbb{R}^{n} $
be a continuously differentiable function and let $\mathbf{y}^* \in \mathbb{R}^n, p_0 \in \mathbb{R}^\ell$ form a solution of the following system of $n$ equations:
\begin{equation}
 F(\mathbf{y}^*, p_0) = 0   \; .
\end{equation}
If the Jacobian matrix $ \nabla_x F(\mathbf{y}^*, p_0) $ is invertible,
then there exists a neighborhood $ U $ of $ p_0 $ and a unique continuously differentiable
function $ \mathbf{y}(\cdot) : U \rightarrow \mathbb{R}^{n} $ such that $ F(\mathbf{y}(p), p) = 0 $ for all $ p \in U $. In addition, the gradient of $ \mathbf{y} $ is given by:
\begin{equation}
    \forall p \in U, \qquad
    \nabla_p \mathbf{y}(p) = - \left( \nabla_{\mathbf{y}} F(\mathbf{y}(p), p) \right)^{-1} \nabla_p F(\mathbf{y}(p), p) \; .
\end{equation}
\end{theorem}

In \cite{fiacco1976sensitivity}, Fiacco investigated under which conditions the KKT equations are regular, in order to
apply the implicit function theorem afterward. To do so, we should ensure that (i) the primal
solution mapping \( \mathbf{y}(p) \) is single-valued (i.e. the problem~\ref{eq:problem} has an unique primal solution) (ii)
the dual solution \( (\lambda(p), \nu(p)) \) is unique and (iii) the active set is locally stable. Fiacco
showed that these three conditions are satisfied if resp. the
constraint qualifications (i) SOSC (ii) LICQ and (iii) SCS hold. In that particular case, the KKT equations~\eqref{eq:Fxp} below rewrite as
a smooth system of nonlinear equations depending only on \( \mathbf{y} \) and \( p \). The implicit function theorem can then be applied to solve the system of equations.

By noting $\mathbf{y} = (x, \lambda, \nu)$, we define
\begin{equation}
    \label{eq:Fxp}
     F(\mathbf{y}, p) =
     \begin{bmatrix}
        \nabla_{x} L(x, \lambda,\nu; p) \\
        c(x; p) \\
        x^\top \nu
    \end{bmatrix} = 0
\end{equation}
The Jacobian w.r.t the solution is given by:
\begin{equation}
  M := \nabla_\mathbf{y} F(\mathbf{y}, p) =
    \begin{bmatrix}
      W & A^\top & - I \\
      A & 0 & 0 \\
      V & 0 & X
    \end{bmatrix} \in \mathbb{R}^{(2n +m) \times (2n + m)}  \; ,
\end{equation}
with the Hessian $W = \nabla_{x x} L(x, \lambda, \nu; p)$, the Jacobian $A = \nabla_{x} c(x; p)$, $X = \text{diag}(x)$ and $V = \text{diag}(\nu)$.

The Jacobian w.r.t the parameters is given by:
\begin{equation}
     N := \nabla_p F(\mathbf{y}, p) =
    \begin{bmatrix}
      \nabla_{xp} L(x, \lambda, \nu; p) \\
      \nabla_{p} c(x; p) \\
      0
    \end{bmatrix} \in \mathbb{R}^{(2n+m) \times \ell}\; .
\end{equation}
Using the IFT, the Jacobian of the solution w.r.t. the parameters
$\nabla_p \mathbf{y}(p)$ is given as solution of the linear system
\begin{equation}
  M \nabla_p \mathbf{y}(p) = - N \; ,
\end{equation}
which translates to the resolution of $\ell$ linear systems. Note that, as these linear systems share the same matrix of equations, the LU factorization only need to be computed once. In general the Jacobian $\nabla_p \mathbf{y}(p)$ is dense.

In practice, most applications do not require evaluating the full Jacobian $\nabla_p \mathbf{y}(p)$.
Following \cite{griewank2008evaluating}, the forward mode computes a \emph{tangent} for
a given \emph{seed direction} $\dot{v} \in \mathbb{R}^\ell$ as
\begin{equation}
  \label{eq:forwarddirection}
  \nabla_p \mathbf{y}(p) \, \dot{v} = - M^{-1} N \dot{v} \;,
\end{equation}
whereas the reverse mode computes a \emph{gradient} for given \emph{weight functionals} $\bar{v} \in \mathbb{R}^n$ as
\begin{equation}
  \label{eq:reversedirection}
  \nabla_p \mathbf{y}(p)^\top \, \bar{v} = - N^\top M^{-\top} \bar{v} \;.
\end{equation}
The computation of the tangent~\eqref{eq:forwarddirection} and the gradient \eqref{eq:reversedirection}
both require the resolution of a single linear system.

%







\section{Main Updates in \texttt{DiffOpt}}
\label{sec:main-updates}

This section summarizes the two additions introduced in this work.

\subsection{Parametric API}
The original interface reports derivatives with respect to low-level coefficients; this is general but inconvenient for most workflows where users naturally reason in terms of named parameters that may appear in multiple expressions. This paper introduces first-class \texttt{Parameter} objects in \texttt{JuMP}, while retaining the low-level API for backtrack compatibility. The parametric layer maintains the mapping from each parameter to all affected terms across MOI bridges and reformulations, so forward and reverse sensitivities can be queried directly with respect to user-declared parameters without altering existing models or solver configurations.

\subsection{Nonconvex Problems}
Prior releases exposed sensitivities for conic and quadratic models. We extend \texttt{DiffOpt.jl} to support smooth, potentially nonconvex NLPs, operating at locally optimal solutions returned by standard nonlinear solvers using the theory discussed in Section \ref{sec:background}. The implementation reuses solver-side linear algebra when available and otherwise assembles the required blocks through \texttt{MathOptInterface.jl}. To ensure numerical robustness at nearly-degenerate solutions, the sensitivity calculation includes light regularization (e.g., small diagonal damping) when needed.

\subsection{Objective Sensitivity}
\label{sec:obj-sens}

In the new version of DiffOpt, we also expose the \emph{objective sensitivity} with respect to explicit parameters. 

If a parameter $p$ enters only as a right-hand side or constant term of a constraint, the objective sensitivity to a small perturbation of $p$ equals the corresponding optimal dual multiplier (under standard KKT regularity and strong duality). This provides an immediate way to read off $\partial J/\partial p$ from the solution. However, when a parameter scales coefficients (e.g., appears multiplied by variables), the sensitivity follows the chain rule:
\[
\frac{\partial f}{\partial p}
\;=\;
\frac{\partial f}{\partial x}\;
\frac{\partial x}{\partial p},
\]
where $\partial x/\partial p$ is obtained from the forward-mode solution sensitivities computed by \texttt{DiffOpt}.

In reverse-mode, the user may query the \emph{parameter cotangent} induced by an \emph{objective seed}. Given a scalar objective perturbation $\bar J$, \texttt{DiffOpt} returns $\bar p \coloneqq \nabla_p J^\top \bar J$, i.e., the first-order change in parameters that would produce the seeded change in the objective. This reverse objective seed is mathematically equivalent to seeding the solution adjoint and propagating through the KKT system; consequently, reverse-mode objective and solution seeds are \emph{mutually exclusive} and cannot be set simultaneously.

\section{Example Use Cases}
We now present three examples of leveraging the features described in Section \ref{sec:main-updates}.
\subsection{Case Study 1: Optimal Power Dispatch Sensitivity}
\label{sec:case-dispatch}

\textbf{Overview.} A classic application is to allocate power generation to meet a demand $d$ at minimum cost, subject to capacity limits. We consider a simplified economic dispatch problem:
\begin{align*}
\min_{g_i \ge 0,\;\phi \ge 0} & \quad \sum_{i=1}^n c_i\,g_i + \sum_{i=1}^n c_{2,i}\,g_i^2  + c_\phi \,\phi \\
\text{s.t.} & \quad \sum_{i=1}^n g_i + \phi \;=\; d, \quad (:\lambda)\\
& \quad 0 \;\le\; g_i \;\le\; G_i, \quad i=1,\dots,n,
\end{align*}
where $g_i$ is the power generated by plant $i$, each with linear unit cost $c_i$, quadratic cost component $c_{2,i}$ and capacity $G_i$, and $\phi$ is unmet demand with penalty $c_\phi$. We treat $d$ (the system demand) as a parameter. Differentiating through this QP quantifies how optimal generation shifts as demand changes.

\noindent\textbf{Implementation in \texttt{DiffOpt.jl}.}
\begin{lstlisting}[language=Julia]
using JuMP, DiffOpt, HiGHS

# 1) Create a diff-capable QP model:
model = DiffOpt.quadratic_diff_model(HiGHS.Optimizer)
set_silent(model)

# 2) Define the parameter for demand:
@variable(model, d in Parameter(100.0)) # e.g., demand=100

# 3) Decision variables
n_plants = 2
c = [20.0, 30.0]   # linear cost of each plant
c2 = [0.2, 0.1]    # quadratic cost of each plant
G = [150.0, 80.0]  # capacity of each plant
@variables(model, begin
    0 <= g[i=1:2] <= G[i]   # generation from each plant
    φ >= 0                  # unmet demand
end)

# 4) Objective: Minimize costs
cφ = 1000.0
@objective(model, Min, sum(c[i]*g[i] for i=1:2) +
    sum(c2[i]*g[i]^2 for i=1:2) + cφ*φ
)

# 5) Demand balance constraint
@constraint(model, sum(g[i] for i=1:n_plants) + φ == d)

# Solve for initial demand=100
optimize!(model)
@show value.(g), value(φ)

# 6) Forward-mode differentiation wrt d
DiffOpt.empty_input_sensitivities!(model)
# Let the direction dd/dd=1.0
DiffOpt.set_forward_parameter(model, d, 1.0)
DiffOpt.forward_differentiate!(model)

# Retrieve dg[i]/dd  and dφ/dd
dg1_dd = DiffOpt.get_forward_variable(model, g[1])
dg2_dd = DiffOpt.get_forward_variable(model, g[2])
dphi_dd = DiffOpt.get_forward_variable(model, φ)
println("dg1/dd = $dg1_dd, dg2/dd = $dg2_dd, dφ/dd = $dphi_dd")
\end{lstlisting}

When $d$ changes by a small amount, \texttt{DiffOpt.forward\_differentiate!} solves a linear system capturing the KKT conditions, revealing how $g_1^*$, $g_2^*$, and $\phi^*$ shift with respect to $d$. These sensitivities are critical for power systems operators to understand how different plants ramp up or down as demand fluctuates.

\begin{figure}[t]
    \centering
    \includegraphics[width=\textwidth]{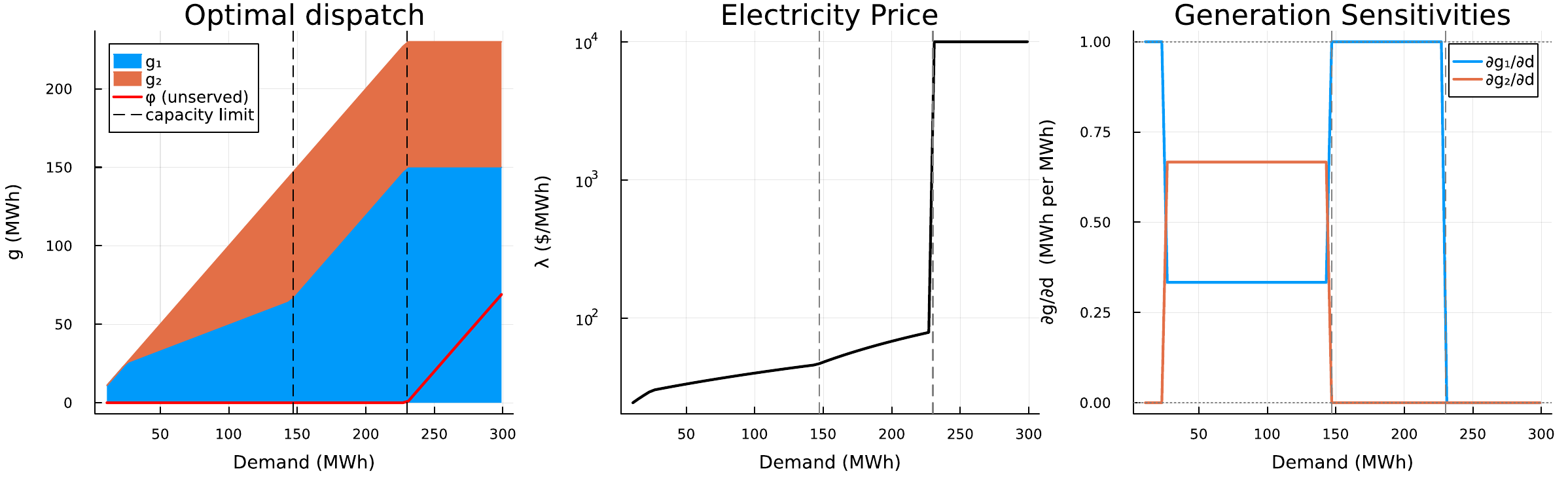}
    \caption{Optimal dispatch (left), marginal electricity price
      $\lambda(d)=\partial J/\partial d$ (center, logarithmic scale),
      and forward sensitivities $\partial g_i/\partial d$ (right) for
      demands ranging from \(\approx 0\) to \SI{300}{\mega\watt\hour}.  The vertical
      dashed lines mark when the individual plant capacities are reached---$g_2$ reaches it's capacity at
      \SI{148}{\mega\watt\hour} and $g_1$ at \SI{230}{\mega\watt\hour}.}
    \label{fig:dispatch_sensitivities}
\end{figure}

Figure~\ref{fig:dispatch_sensitivities} illustrates how optimal generation, the marginal
price~$\lambda(d)$, and their sensitivities vary as demand~$d$ changes.  At low demand
only generator~1 is marginal, so $\partial g_1/\partial d = 1$ while
$\partial g_2/\partial d = 0$; every forecast error is absorbed by the cheaper
unit and the price rises along its quadratic cost curve.
Once demand reaches \SI{27}{MWh}, generator~2 also becomes marginal and the
incremental load is shared, with generator~2 taking twice as much:
$\partial g_2/\partial d = \tfrac{2}{3}$ and
$\partial g_1/\partial d = \tfrac{1}{3}$.  At $d = \SI{148}{MWh}$,
generator~2 reaches its capacity and generator~1 again absorbs all further
changes, restoring $\partial g_1/\partial d = 1$.  Approaching the
\SI{230}{MWh} combined limit, the sensitivities plunge toward zero because any
extra megawatt would be curtailed through the slack variable~$\phi$.
Interpreting the sensitivities reveals
(i) which unit will ramp next allowing operators to communicate preemptively with the companies that manage those generators, (ii) and how much flexibility remains---an important factor for system reliability.  This insight
is even more valuable in realistic, network-constrained dispatch problems—where
generators and loads occupy different buses—because identifying which units
can absorb a local fluctuation is far less obvious without the guidance
provided by these sensitivities.

\bigskip

\subsection{Case Study 2: Mean--Variance Portfolio Optimization}
\label{sec:case-portfolio}

\textbf{Overview.} Consider the Markowitz portfolio selection problem, which allocates weights $x \in \mathbb{R}^n$ to $n$ assets so as to maximize returns subject to a variance limit $v_{\max}$:
\[
\max_{x} \quad \mu^\top x
\quad\text{s.t.}\quad
x^\top \Sigma x \;\le\; v_{\max}, \quad
\mathbf{1}^\top x = 1,\quad
x \succeq 0,
\]
where $\mu$ is the vector of expected returns, $\Sigma$ is the covariance matrix, and $x$ must sum to 1 (fully invest the budget). An efficient conic version of this problem casts the variance limit as a second order cone constraint:

\[
\| \Sigma^{1/2} x \|_{2} \;\le\; \sigma_{\max}
\]
where $\Sigma^{1/2}$ is the Cholesky factorization of the covariance matrix and $\sigma_{\max}$ is the standard deviation limit.

Practitioners often care about an \emph{out-of-sample performance metric} $L(x)$ evaluated on test data or scenarios that differ from those used to form $\mu$ and $\Sigma$. To assess the impact of the risk profile in the performance evaluation, one can compute:
\[
\frac{dL}{d\,\sigma_{\max}} \;=\;
\underbrace{\frac{\partial L}{\partial x}}_{\text{(1) decision impact}}\;
\cdot\;
\underbrace{\frac{\partial x^*}{\partial \sigma_{\max}}}_{\text{(2) from DiffOpt.jl}},
\]
where $x^*(\sigma_{\max})$ is the portfolio that solves the conic Markowitz problem under a given risk limit.

\noindent\textbf{Reverse Mode Implementation in \texttt{DiffOpt.jl}.}
\begin{lstlisting}[language=Julia]
using JuMP, DiffOpt, COSMO, LinearAlgebra

# 1) Create a diff-capable conic model
model = DiffOpt.conic_diff_model(COSMO.Optimizer)
set_silent(model)

# 2) Parameter: standard deviation limit (σ_max)
@variable(model, σ_max in Parameter(0.04)) # e.g. limit = 0.04

# 3) Decision variables: x (portfolio weights)
n = 3
@variable(model, x[1:n] >= 0)
@constraint(model, sum(x) == 1)

# 4) Covariance & returns (toy data)
Sigma = [0.002  0.0005  0.001;
         0.0005 0.003   0.0002;
         0.001  0.0002  0.0025]
mu = [0.05, 0.08, 0.12]

# Perform a Cholesky factorization of Sigma
# (For numerical stability, ensure Sigma is PSD)
L = cholesky(Symmetric(Sigma)).L

# 5) Objective: maximize mu' * x
@objective(model, Max, sum(mu[i] * x[i] for i in 1:n))

# 6) Conic constraint for the variance limit:
#     || L*x || <= σ_max
# This enforces x' * Sigma * x <= (σ_max)^2
# by requiring  norm(L*x, 2) <= σ_max
@variable(model, t >= 0)
@constraint(model, [t; L * x] in SecondOrderCone())
@constraint(model, t <= σ_max)

# Solve once for initial σ_max
optimize!(model)

@show value.(x)  # display weights for the initial σ_max

# Clear any previously set AD seeds, etc.
DiffOpt.empty_input_sensitivities!(model)

# 7) Reverse-mode: set the reverse-variable (adjoint) to dL/dx
# we set it random just for demostration purposes
DiffOpt.set_reverse_variable.(model, x, rand(n))

# 8) Differentiate in reverse mode
DiffOpt.reverse_differentiate!(model)

# Retrieve sensitivity: dL/dσ_max
dL_dσmax = DiffOpt.get_reverse_parameter(model, σ_max)
println("dL/dσ_max = ", dL_dσmax)
\end{lstlisting}

\begin{figure}[H]
    \centering
    \includegraphics[width=\textwidth]{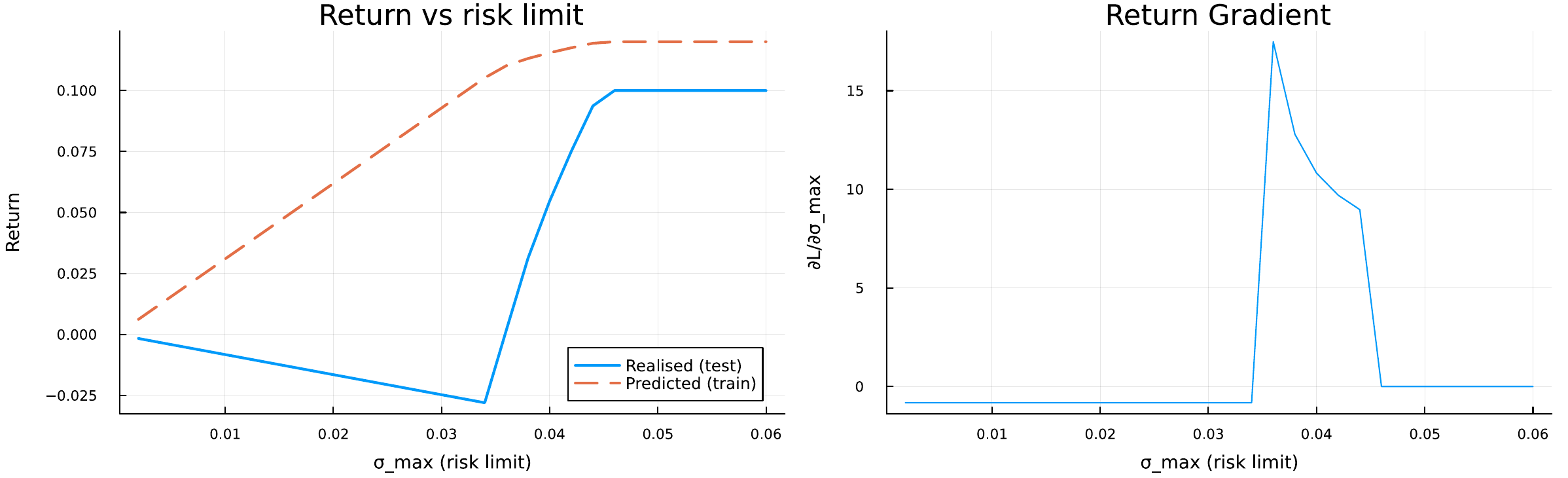}
    \caption{Impact of the risk limit \(\sigma_{\max}\) on Markowitz
      portfolios.  \textbf{Left:} predicted in-sample return versus
      realized out-of-sample return.  \textbf{Right:} the
      out-of-sample loss \(L(x)\) together with the absolute gradient
      \(|\partial L/\partial\sigma_{\max}|\) obtained from
      \texttt{DiffOpt.jl}.  The gradient tells the practitioner which
      way—and how aggressively—to adjust \(\sigma_{\max}\) to reduce
      forecast error; its value is computed in one reverse-mode call
      without re-solving the optimization for perturbed risk limits.}
    \label{fig:portfolio_risk_sweep}
\end{figure}

This conic-program example highlights how one can treat $\sigma_{\max}$ (the risk tolerance) as a tunable parameter within a larger \emph{learning or validation} pipeline. By differentiating the Markowitz problem in reverse mode, we see how a small change in $\sigma_{\max}$ propagates to the final portfolio $x^*$ and, in turn, how that shift affects an external performance metric $L(x)$. Figure \ref{fig:portfolio_risk_sweep} shows how loosening or tightening the risk limit alters the
bias (difference between predicted and realized return) and how the
DiffOpt gradient indicates the direction that decreases the
out-of-sample loss.

This approach effectively unifies risk modeling and data-driven learning: rather than manually searching over volatility limits, a gradient-based method can now \emph{automatically} adjust $\sigma_{\max}$ to optimize out-of-sample performance in an end-to-end differentiable framework.

\bigskip

\subsection{Case Study 3: Nonlinear Robot Inverse Kinematics}
\label{sec:case-robot-ik}

\textbf{Overview.} Inverse Kinematics (IK) computes joint angles that place a robot’s end-effector at a desired target $(x_t,y_t)$. For a 2-link planar arm with joint angles $\theta_1,\theta_2$, the end-effector position is:
\[
f(\theta_1,\theta_2) = \bigl(\ell_1\cos(\theta_1) + \ell_2\cos(\theta_1+\theta_2),\,\,
\ell_1\sin(\theta_1) + \ell_2\sin(\theta_1+\theta_2)\bigr).
\]
We can solve an NLP:
\[
\min_{\theta_1,\theta_2} \;\; (\theta_1^2 + \theta_2^2),
\quad\text{s.t.}\quad f(\theta_1,\theta_2) = (x_t,y_t).
\]
Treat $(x_t,y_t)$ as parameters. Once solved, we differentiate w.r.t. $(x_t,y_t)$ to find how small changes in the target location alter the optimal angles---the \emph{differential kinematics}.

\begin{figure}[H]
\centering
\includegraphics[width=0.8\linewidth]{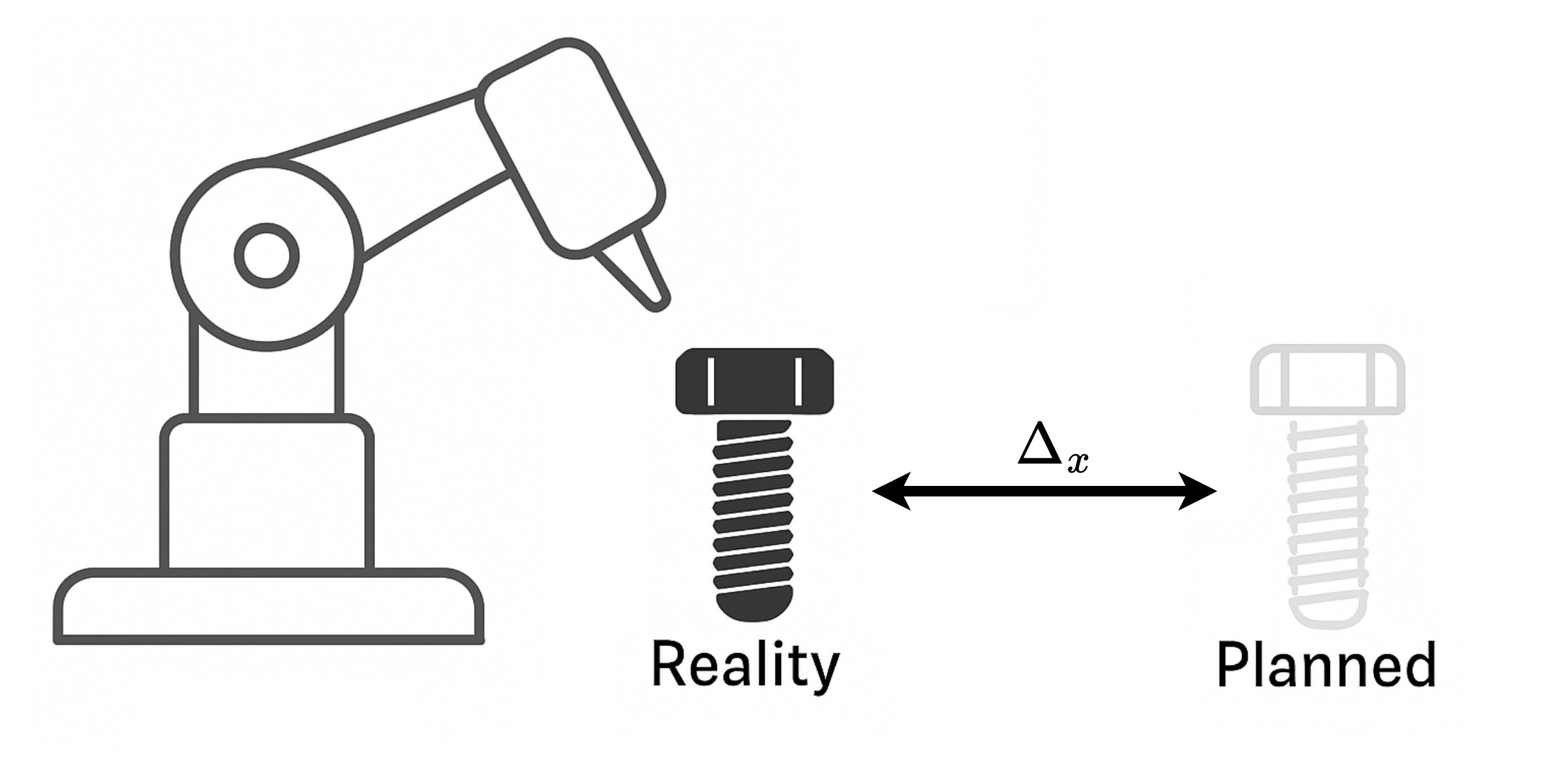}
\caption{Planned vs. Observed reality for a robot arm.}
\label{fig:ik_plan}
\end{figure}

\noindent\textbf{Implementation in \texttt{DiffOpt.jl}.}
\begin{lstlisting}[language=Julia]
using JuMP, DiffOpt, Ipopt

model = DiffOpt.nonlinear_diff_model(Ipopt.Optimizer)
set_silent(model)

@variable(model, x_t in Parameter(1.0))  # target x
@variable(model, y_t in Parameter(1.0))  # target y

# Two joint angles, unconstrained
@variable(model, θ1)
@variable(model, θ2)

# Objective: minimize squared angles
@objective(model, Min, θ1^2 + θ2^2)

# Nonlinear constraints for 2-link kinematics
l1, l2 = 1.0, 1.0
@constraint(model, cos(θ1)*l1 + cos(θ1+θ2)*l2 == x_t)
@constraint(model, sin(θ1)*l1 + sin(θ1+θ2)*l2 == y_t)

optimize!(model)
@show value(θ1), value(θ2)

# Forward-mode: partial derivatives w.r.t. x_t
DiffOpt.empty_input_sensitivities!(model)
DiffOpt.set_forward_parameter(model, x_t, 1.0)
DiffOpt.forward_differentiate!(model)
dθ1_dx = DiffOpt.get_forward_variable(model, θ1)
dθ2_dx = DiffOpt.get_forward_variable(model, θ2)
println("dθ1/dx_t = $dθ1_dx,  dθ2/dx_t = $dθ2_dx")
\end{lstlisting}

By repeating for \texttt{y\_t}, one obtains $\partial\theta/\partial y_t$. These derivatives match the implicit function theorem applied to the KKT conditions.

\begin{figure}[t]
    \centering
    \includegraphics[width=\linewidth]{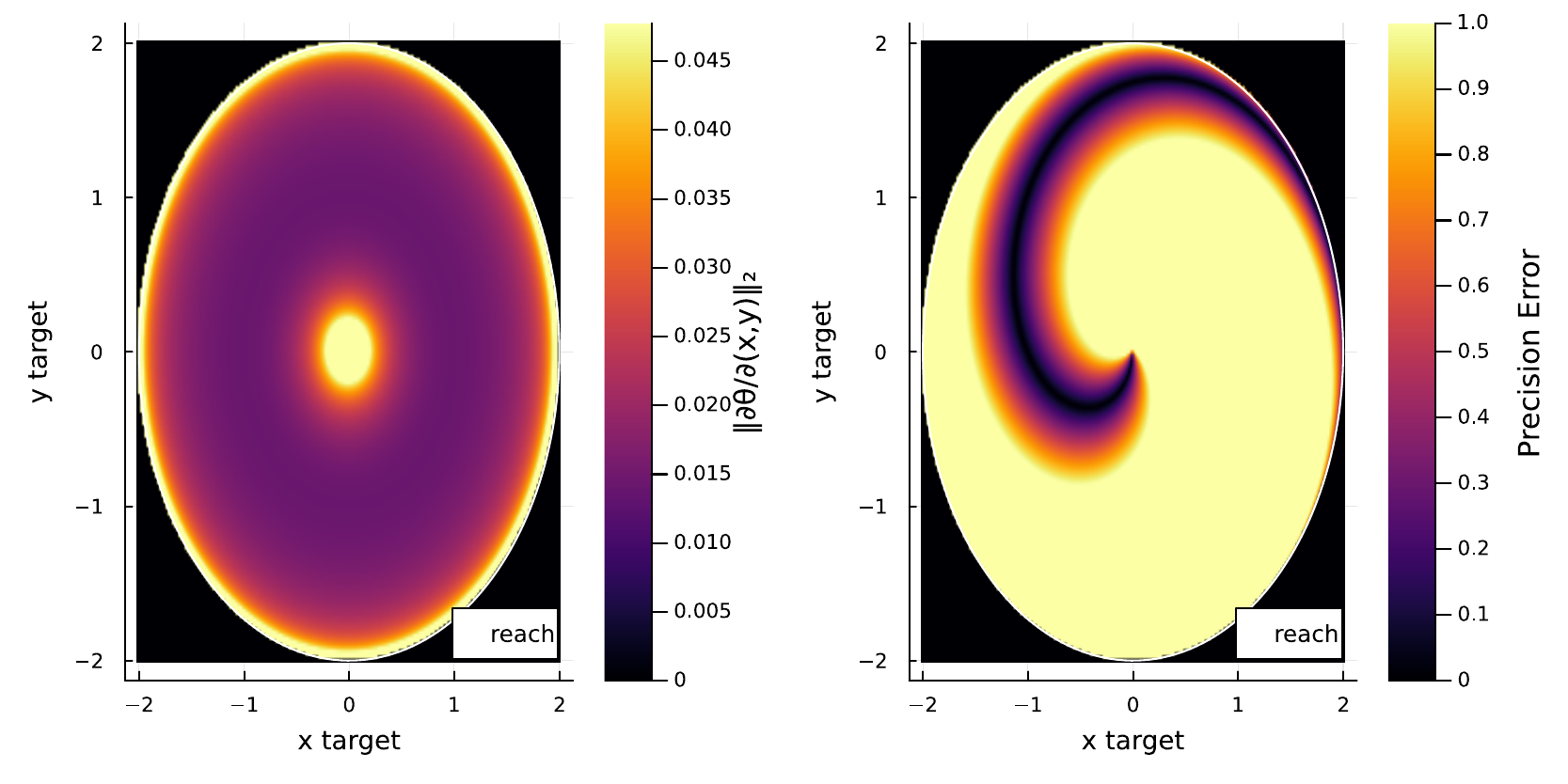}
    \caption{Left figure shows the spectral-norm heat-map
      \(\bigl\lVert\partial\boldsymbol{\theta}/\partial(x,y)\bigr\rVert_2\)
      for a two-link arm - Bright rings mark near-singular poses. Right figure shows the normalized precision error of the first order approximation derived from calculated sensitivities.}
    \label{fig:ik_heat}
  \hfill
\end{figure}

A small $\Delta x_t$ implies a linearized change in $(\theta_1,\theta_2)$---making it easy to do small robot re-targeting without re-solving a brand-new IK from scratch. Figure \ref{fig:ik_heat} shows the magnitude and precision of the sensitivities for this problem using the proposed formulation. It is important for both operator and designer of such robot arm to understand the robustness and stability of the robot in different regions.

\section{Conclusion}
\label{sec:conclusion}
This work unifies modeling and differentiation for constrained optimization in Julia. We introduced a parameter-centric API that exposes forward- and reverse-mode \emph{solution} and \emph{objective} sensitivities directly with respect to named parameters, and extended \texttt{DiffOpt.jl} beyond convex programs to smooth, potentially nonconvex NLPs. By operating natively through \texttt{JuMP}/\texttt{MOI}, the framework leverages existing solvers and transformations while keeping user models unchanged. Case studies in energy systems, portfolio optimization, and nonlinear robotics illustrate that these capabilities make differentiable optimization a practical, scalable component of learning, calibration, and design workflows.

\paragraph{Limitations and future work.}
Our approach computes local derivatives around KKT-regular solutions and, as such, inherits the usual issues of degeneracy and active-set changes. Extending robustness to nonsmooth regimes, mixed-integer formulations, and broader conic–nonlinear composites remains an open direction. On the systems side, we see opportunities in tighter reuse of solver factorizations, batched/multi-parameter differentiation, second-order (Hessian–vector) products, richer diagnostics for stability, and deeper integrations with ML toolchains. We hope these additions encourage routine use of differentiable optimization at scale and catalyze further advances across theory, algorithms, and applications.

\bibliographystyle{unsrt}
\bibliography{bibliography}

\end{document}